\documentclass{article}

\usepackage[preprint]{neurips_2023}
\usepackage{natbib}
\setcitestyle{numbers,square,comma}

\usepackage[utf8]{inputenc} 
\usepackage[T1]{fontenc}    
\usepackage{hyperref}       
\usepackage{url}            
\usepackage{booktabs}       
\usepackage{amsfonts}       
\usepackage{nicefrac}       
\usepackage{microtype}      
\usepackage{xcolor}         
\usepackage{graphicx}
\usepackage{wrapfig}
\usepackage{multirow}
\usepackage{epsfig}
\usepackage{amsmath}
\usepackage{amssymb}
\usepackage{marvosym}
\usepackage{color}
\usepackage{threeparttable}
\usepackage{algorithm}
\usepackage{algpseudocode}
\usepackage{setspace}
\usepackage{amsthm}
\usepackage[hang,flushmargin]{footmisc}
\usepackage{subcaption}
\usepackage{float}
\usepackage{natbib}

\newcommand{\boldres}[1]{{\textbf{\textcolor{red}{#1}}}}
\newcommand{\secondres}[1]{{\underline{\textcolor{blue}{#1}}}}

\title{CVTN: Cross Variable and Temporal Integration for Time Series Forecasting}

\author{%
  Han Zhou\\
  Zhejiang University\\
  Ningbo Institute of Digital Twin, Eastern Institute of Technology, Ningbo \\
  \texttt{zhouhan0315@zju.edu.cn} \\
  \AND
  Yuntian Chen\thanks{* Corresponding author} \\
  Ningbo Institute of Digital Twin, Eastern Institute of Technology, Ningbo \\
  \texttt{ychen@eitech.edu.cn} 
}

\begin{document}
\maketitle

\begin{abstract}
  In multivariate time series forecasting, the Transformer architecture encounters two significant challenges: effectively mining features from historical sequences and avoiding overfitting during the learning of temporal dependencies. To tackle these challenges, this paper deconstructs time series forecasting into the learning of historical sequences and prediction sequences, introducing the Cross-Variable and Time Network (CVTN). This unique method divides multivariate time series forecasting into two phases: cross-variable learning for effectively mining features from historical sequences, and cross-time learning to capture the temporal dependencies of prediction sequences. Separating these two phases helps avoid the impact of overfitting in cross-time learning on cross-variable learning. Extensive experiments on various real-world datasets have confirmed its state-of-the-art (SOTA) performance. CVTN emphasizes three key dimensions in time series forecasting: the short-term and long-term nature of time series (locality and longevity), feature mining from both historical and prediction sequences, and the integration of cross-variable and cross-time learning. This approach not only advances the current state of time series forecasting but also provides a more comprehensive framework for future research in this field.
\end{abstract}

\section{Introduction}
As artificial intelligence technologies advance unceasingly, the role of time series prediction in key sectors such as energy management\cite{gao2023adaptive}, meteorology\cite{meenal2022weather}, finance\cite{lopez2023can}, and sensor networks\cite{mejia2020prediction} has grown increasingly critical. Long-term Time Series Forecasting (LTSF), involving projections far into the future, is crucial for strategic planning, providing significant reference value. 

The inherent constraints of conventional statistical techniques in handling complex time series prediction tasks have spurred an increasing interest among data scientists in the application of deep learning methodologies for forecasting. Throughout years of evolution and competitive advancements, the Time-Series Forecasting Transformer\cite{vaswani2017attention} (TSFT), distinguished by its superior sequence modeling abilities and scalability, has become widely adopted for long-term time series forecasting.

The TSFT architecture generally consists of an encoder and a decoder. The encoder analyzes the input historical time series, enriched with positional encoding to incorporate temporal context, and employs self-attention mechanisms and feed-forward neural networks (FFN) to extract pivotal features of the series. The decoder, aimed at generating the predictive sequence, initiates with either a start symbol or a predetermined historical sequence. It progressively builds the output sequence, employing cross-attention mechanisms to integrate the encoder’s output vectors for enhanced prediction accuracy.

Nonetheless, TSFT has encountered skepticism from researchers. Prior research\cite{gao2023client} has revealed that TSFT's efficacy remains relatively unaffected even when portions of the historical sequence are obscured, raising questions about its ability to derive genuinely significant information from these sequences. To counter this, certain TSFT models utilizing cross-variable transformers have demonstrated advancements in long-term forecasting\cite{gao2023client, liu2023itransformer, zhang2022crossformer}. These models significantly elevate performance, particularly in datasets characterized by multi-variable interdependencies.

Cross-variable TSFT effectively extracts pertinent information from historical sequences by identifying multi-variable dependencies. Nonetheless, there is significant potential for enhancement in grasping temporal dynamics and delineating temporal correlations within prediction sequences. The Crossformer introduces a two-stage attention (TSA) layer to discern cross-temporal dependencies and interdependencies across different dimensional segments of the series. This approach combines temporal and variable dependency learning, potentially causing interference between these two modes of information extraction. Consequently, its performance often falls short of linear models in many datasets. In light of this, this paper advocates for the segregation of cross-variable and cross-temporal learning to attain higher prediction accuracy across various datasets.

Our examination of the Cross-Temporal Transformer model across diverse datasets indicates that Transformers segmenting tokens temporally tend to overfit, hindering the model's capacity to learn genuinely effective feature representations from historical data. The study further discerns that its predictive prowess predominantly originates from the encoder's proficiency in understanding the temporal dynamics of the target sequence. This underscores the imperative to adeptly mine historical sequence data while employing cross-temporal encoding networks to accurately model the temporal relationships in the predictive sequence.

Moreover, while research in long-term time series forecasting frequently emphasizes the extended duration of sequences, it is observed that time series often display marked variations over longer time spans. As time progresses, the statistical attributes of the series, such as mean and variance, tend to evolve. Nonetheless, within a localized context, time series commonly show certain patterns of regularity. For instance, in electricity usage forecasting, variations in usage over a short range usually occur gradually, maintaining relatively stable statistical characteristics with more consistent recognition patterns. Informed by this observation, the cross-temporal model introduced in this paper leverages a CNN-based structure, highlighting the local dependencies within the target sequence.

In conclusion, drawing from the aforementioned analyses, we introduce a dual-phase deep learning network architecture. The initial phase, termed the Cross-Variable Encoder (CVE), is tailored to discern inter-variable dependencies, effectively extracting information from historical sequences. Subsequently, the Cross-Temporal Encoder (CTE), constituting the second phase, assimilates both the original input and the CVE output, focusing on learning cross-temporal dependencies. This approach addresses the limitations in cross-temporal feature learning inherent in the first stage and delineates the temporal relationships within predictive sequences. Furthermore, by distinctively segregating cross-variable and cross-temporal learning, our model significantly reduces the risk of overfitting commonly associated with cross-temporal learning.
This research makes three significant contributions to the field of time series forecasting:
\begin{itemize}
\item Introducing CVTN (Cross Variable and Temporal Network): This paper unveils CVTN, a novel deep learning network that distinctively isolates and then combines cross-variable with cross-temporal feature extraction. Within CVTN, CVE (Cross-Variable Encoder) efficiently mines data from historical sequences, while CTE (Cross-Temporal Encoder) overcomes its limitations in dynamic temporal feature learning, focusing on the associative relationships in predictive sequences. CVTN's architecture, by separating these two learning processes, effectively reduces overfitting in temporal feature learning.
\item Key Elements in Time Series Forecasting:
a) Overfitting in Cross-Temporal Learning: A significant factor behind models' failure to extract meaningful information from historical data is overfitting in cross-temporal learning. Our experiments show that Transformer models are not fully leveraging historical sequence data, deriving most of their efficacy from the associative relationships within target sequences.
b) Importance of Local Dependencies: In time series forecasting, prioritizing local dependencies, particularly the temporal relationships between predictive sequences, is crucial. This often-neglected aspect is vital for precise forecasting.
\item The CVE and CTE components of CVTN are designed to be lightweight, achieving state-of-the-art (SOTA) performance on multiple real-world datasets with minimal computational overhead. The experiments conducted robustly demonstrate CVTN's high accuracy and resilience, making it a significant advancement in the field of time series forecasting.
\end{itemize}

\section{Related Work}
The Transformer model, renowned for its outstanding performance in various fields such as natural language processing, speech recognition, and computer vision, has been adapted for time series forecasting through a variety of variants to enhance its self-attention mechanism. These adaptations mainly aim to learn long-term dependencies using cross-Temporal attention mechanisms and to optimize computational efficiency.

LogTrans\cite{li2019enhancing} implements a convolutional self-attention layer with a LogSparse design. This unique approach is adept at capturing local information and simultaneously reduces spatial complexity. Other models like Informer\cite{Zhou_Zhang_Peng_Zhang_Li_Xiong_Zhang_2022} and Autoformer\cite{wu2021autoformer} have innovated by replacing the traditional self-attention mechanism, thereby lowering the computational complexity to \(O(L \log L)\). Pyraformer\cite{liu2021pyraformer} stands out by integrating pyramid attention modules that connect across and within scales, achieving linear complexity in the process.

Further advancements feature models like Autoformer, FEDformer\cite{zhou2022fedformer}, and ETSformer\cite{woo2022etsformer} that integrate TSFT with seasonal trend decomposition and signal processing techniques, including Fourier analysis, within their attention frameworks. This integration not only boosts the interpretability of these models but also efficiently captures seasonal trends within the data.

Addressing the issue of stability in predictions, especially in non-stationary contexts, some Transformer models have incorporated stabilization modules and De-stationary Attention into the standard Transformer framework\cite{liu2022non}. This integration helps in stabilizing the model's predictions while avoiding the pitfalls of excessive stabilization, which can lead to loss of important data variability.

In the realm of long-term multivariate prediction, recent developments in cross-variable Transformer models have shown significant promise. Models like Client\cite{gao2023client} and iTransformer\cite{liu2023itransformer} have improved performance in long-term multivariate forecasting by replacing cross-Temporal Transformers with cross-variable ones. Additionally, Crossformer\cite{zhang2022crossformer} employs a two-stage attention (TSA) layer, aiming to capture dependencies both over time and across different dimensional segments of the series. However, there is room for improvement in models like Crossformer in terms of their performance on various benchmark datasets.

\section{Cross Variable and Temporal Network}
\subsection{Cross-Variable Encoder}
\begin{figure}[!htbp]
  \vspace{-10pt}
    \includegraphics[width=\columnwidth]{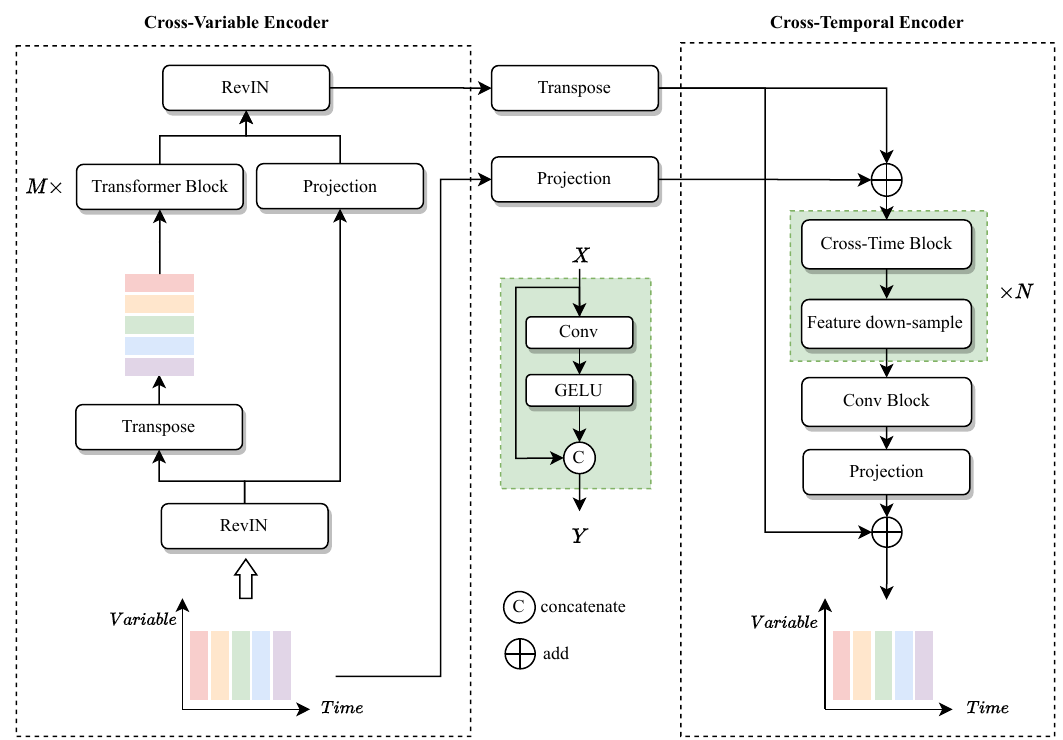}
    \centering
    \caption{{The CVTN architecture is strategically bifurcated into two key components. On the left, CVE adopts a Client\cite{gao2023client} architecture, leveraging the Cross-Variable Transformer to effectively delineate dependencies among variables. In contrast, on the right, CTE utilizes a Convolutional Neural Network (CNN) architecture, meticulously designed to decode cross-temporal dependencies.}} 
    \label{fig:cvtn}
  \end{figure}

As shown in Figure \ref{fig:cvtn}, CVE adopts a client-based architecture as its core to learn dependencies among variables. This architecture differs from the standard Transformer encoder in that CVE integrates an additional linear layer, specifically designed for extracting trends from sequential data. The hallmark of CVE lies in its novel approach to token partitioning. Unlike traditional methods, CVE segments tokens along the variable dimension, with each token representing different temporal instances of the same variable. This is achieved by transposing the input data. The process is illustrated as follows:
\begin{equation}
  \mathbf{V}^0 = \operatorname{Transpose}(\mathbf{X}_{\text{enc}})
\end{equation}
\begin{equation}
  \mathbf{V}^{(m+1)} = \operatorname{Transformer Block}(\mathbf{V}^{m}), \quad m \in \{0, 1, \ldots, M-1\}
\end{equation}
\begin{equation}
  \mathbf{Z}_{\text{CVE}} = \operatorname{Projection}(\mathbf{V}^{M})
\end{equation}

The operational sequence commences by converting the input data \( \mathbf{X}_{\text{enc}} \) into its transposed counterpart \( \mathbf{V}^0 \), where \( \mathbf{V} \) denotes a matrix with \( D \) embedded tokens, each of dimension \( S \). In this context, \( \mathbf{V}^0 \in \mathbb{R}^{D \times S} \) represents the preliminary embedded form of the input. The superscript in \( \mathbf{V}^{(m+1)} \) indicates the layer index within the progression of transformations.

Each subsequent layer \( \mathbf{V}^{(m+1)} \) is generated by applying a \textit{TransformerBlock} to the output of the previous layer \( \mathbf{V}^{m} \), iterating this process for \( m \) in the set \{0, 1, \ldots, M-1\}. The \textit{Transformer Block} typically consists of self-attention mechanisms and a shared FFN, allowing the variate tokens within \( \mathbf{V} \) to interact and be processed independently at each layer. This iterative process enriches the representation of the data by capturing complex dependencies and patterns.

Finally, the Projection operation transforms the output of the last Transformer layer \( \mathbf{V}^{M} \) into a new representation \( \mathbf{Z}_{\text{CVE}} \), where \( \mathbf{Z}_{\text{CVE}} \in \mathbb{R}^{O \times D} \). This operation is implemented by a linear layer.

CVE channels the extracted features into a projection layer to generate first-stage predictions, deliberately omitting a decoder. This approach stems from the decoder's inherent assumption of future sequence invisibility, which overlooks the constraining influence of future sequences on historical data. Additionally, the Transformer module within CVE operates predominantly as a feature extractor rather than a sequence generator, given the absence of temporal interrelations among different variables.

To address the issue of distribution shift, CVE employs a reversible instance normalization (RevIN)\cite{kim2021reversible} module. This module, characterized by its symmetrical structure, can remove and restore the statistical information of time series instances, thereby enhancing the model's stability during the prediction process.

\subsection{Cross-Temporal Encoder}
\begin{algorithm*}[!htbp]
  \setstretch{1.5}
  \caption{CVTN}\label{algo:cvt}
  \begin{algorithmic}[1]
    \Require
    Input time series: $\mathbf{X} \in \mathbb{R}^{L \times C}$; Input Length $L$; Number of Variables $C$; Predict Length $O$; Number of Encoder Layers $N$; Number of CTE Layers $M$.
    \State \textit{\textbf{CVE Stage:}}
    \State $\mathbf{X}^{\prime} = \texttt{RevIN}(\mathbf{X},encode)$
    \Comment{$\mathbf{X}^{\prime} \in \mathbb{R}^{L \times C}$}

    \State $\mathbf{V}^0 = \operatorname{Transpose}(\mathbf{X}^{\prime})$
    \Comment{$\mathbf{V}^0 \in \mathbb{R}^{C \times L}$}
    \State \textbf{for} $m = 0$ \textbf{to} $M-1$ \textbf{do}
    \State $\quad \mathbf{V}^{(m+1)} = \operatorname{TransformerBlock}(\mathbf{V}^m)$
    \State \textbf{end for} \Comment{$\mathbf{V}^{m+1} \in \mathbb{R}^{C \times L}$}
    \State $\mathbf{Z}^{\prime}_{\text{CVE}} = \operatorname{Projection}(\mathbf{V}^M)$ \Comment{$\mathbf{Z}^{\prime}_{\text{CVE}} \in \mathbb{R}^{O \times C}$}
    \State $\mathbf{Z}_{\text{CVE}} = \texttt{RevIN}(\mathbf{Z}^{\prime}_{\text{CVE}},decode)$ \Comment{$\mathbf{Z}_{\text{CVE}} \in \mathbb{R}^{O \times C}$}
    \State \textit{\textbf{CTE Stage:}}
    \State $\mathbf{T}^0 = \mathbf{Z}_{\text{proj}} + \mathbf{Z}_{\text{CVE}}$ \Comment{$\mathbf{T}^0 \in \mathbb{R}^{O \times C}$}
    \State \textbf{for} $n = 0$ \textbf{to} $N-1$ \textbf{do}
    \State $\quad \mathbf{T}^{n+1} = \operatorname{FDS}(\operatorname{CrossTimeBlock}(\mathbf{T}^n))$
    \State \textbf{end for}  \Comment{$\mathbf{T}^N \in \mathbb{R}^{{(C + (r * N) / 2)} \times O}$}
    \State $\mathbf{Y} = \mathbf{Z}_{\text{CVE}} \oplus \operatorname{Projection}(\mathbf{T}^N)$ \Comment{$\mathbf{Y} \in \mathbb{R}^{O \times C}$}
    \State \textbf{return} $\mathbf{Y}$
    \Comment{Return the final prediction result}
  \end{algorithmic}
\end{algorithm*}
CTE plays a pivotal role in modeling the temporal dependencies within a sequence. It processes inputs comprising the projected outputs of the original target sequence melded with the results from CVE. This amalgamation of data enables the CTE to adeptly capture the predictive sequence's temporal dependencies, effectively addressing the CVE stage's limitations in discerning dynamic temporal characteristics.

The Cross-Temporal Block consists of an convolutional layer and employs a concatenation operation to ensure that no information is lost from the input. To avoid performance degradation and the risk of overfitting due to an excess of features, we employ point-wise convolutions to construct a Feature down-sample (FDS) module, which halves the input features. The output of CTE is then combined with the output of CVE through an additive fusion process to optimize the residual between CTE and the predictive sequence. CTE is simply expressed as:

\begin{equation}
  \mathbf{T}^0 = \mathbf{Z}_{\text{proj}} \oplus \mathbf{Z}_{\text{CVE}}
\end{equation}
\begin{equation}
  \mathbf{T}^{n + 1} = \operatorname{FDS}(\operatorname{CrossTimeBlock}(\mathbf{T}^{n})), \quad \text{for } n \in \{0, 1, \ldots, N - 1\}
\end{equation}
\begin{equation}
  \mathbf{Y} = \mathbf{Z}_{\text{CVE}} \oplus \operatorname{Projection}(\mathbf{T}^{N})
\end{equation}

where \( \mathbf{T}^0 \) denotes the initial input state, formed by the addition of \( \mathbf{Z}_{\text{proj}} \) and \( \mathbf{Z}_{\text{CVE}} \), where \( \mathbf{T}^0 \) resides in the space \( \mathbb{R}^{O \times D} \). This signifies that \( \mathbf{T}^0 \) contains \( O \) embedded tokens, each of dimension \( D \), capturing the combined information from the projected target sequence and the output of CVE. \( n \) indicates the layer index in the sequence of transformations, iterating from 0 to \( N - 1 \). FDS and the CrossTimeBlock interactively refine the temporal features in each layer. Finally, the cumulative output of this sequential operation, \( \mathbf{T}^{N} \), is combined with the CVE's output.

\subsection{Cross Variable and Temporal Network}
CVTN employs a two-stage training process. The first stage is cross-variable learning, which focuses on learning the dependencies between variables. This stage is instrumental in capturing the relationships and interactions across different variables in the dataset. The second stage is cross-Temporal learning, which aims to supplement the shortcomings in temporal information mining from the first stage. Additionally, it leverages the most beneficial aspect of temporal models, namely the modeling of temporal relationships in the predictive sequence. The complete algorithm of CVTN is illustrated in Algorithm  \ref{algo:cvt}.

The training process of CVTN is conducted in separate stages to prevent the interference of cross-time learning with cross-variable learning. During the cross-variable learning phase, only CVE is updated using gradients, while CTE does not undergo any gradient updating. The output from this phase is compared with actual labels to calculate loss and generate an initial prediction sequence. In the subsequent cross-time learning phase, the CVE remains unchanged, and only the CTE is updated. This method is specifically designed to capture the temporal dependencies among the prediction sequences.

\begin{figure}[!htbp]
  \centering
  \begin{subfigure}{.5\textwidth}
      \centering
      \includegraphics[width=0.9\linewidth]{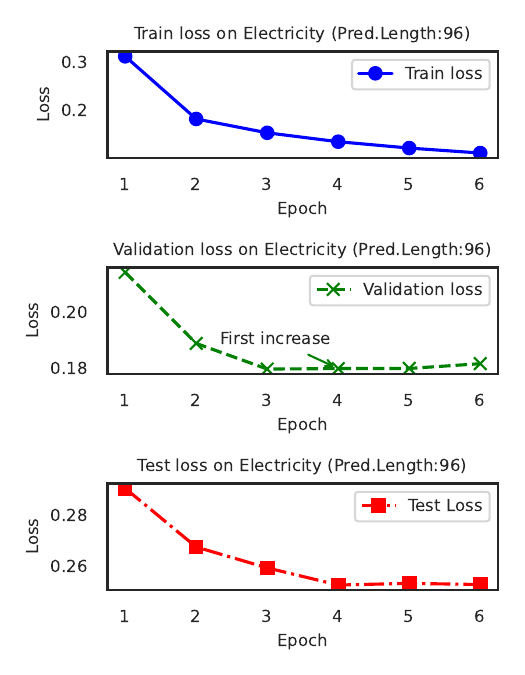}
      
  \end{subfigure}%
  \begin{subfigure}{.5\textwidth}
      \centering
      \includegraphics[width=0.9\linewidth]{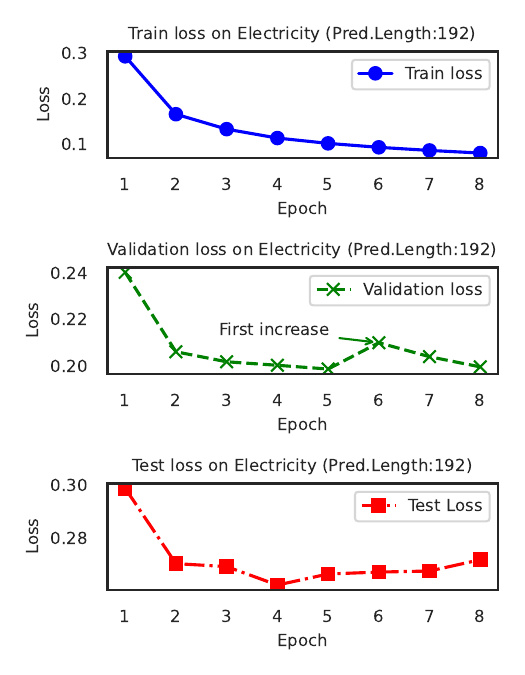}
  \end{subfigure}
  \begin{subfigure}{.5\textwidth}
    \centering
    \includegraphics[width=0.9\linewidth]{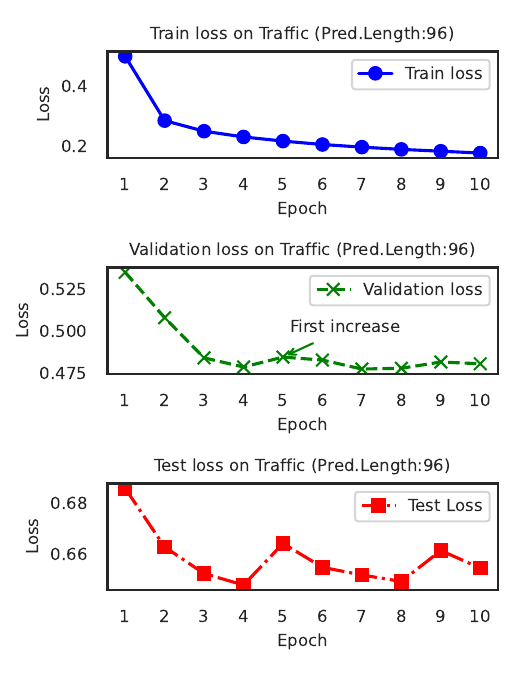}
  \end{subfigure}%
  \begin{subfigure}{.5\textwidth}
      \centering
      \includegraphics[width=0.9\linewidth]{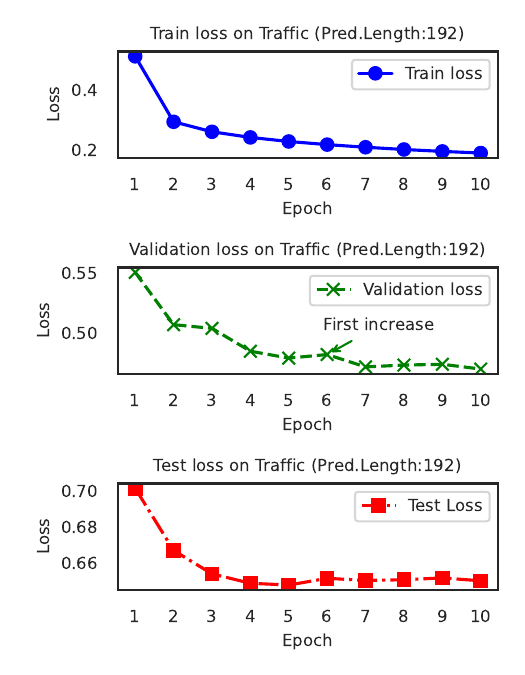}\
  \end{subfigure}
  
  \caption{Observation of the model's loss trend on the Electricity and Traffic datasets. Training was fixed for 10 epochs with an early stopping tolerance of 3. Training was terminated upon exceeding this tolerance level.}
  \label{fig:TSFT}
\end{figure}

\begin{figure}[h]
  \centering
  \begin{subfigure}{.5\textwidth}
      \centering
      \includegraphics[width=1.0\linewidth]{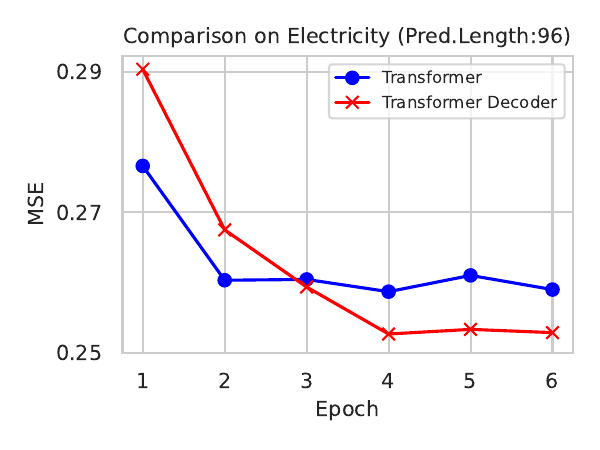}
  \end{subfigure}%
  \begin{subfigure}{.5\textwidth}
      \centering
      \includegraphics[width=1.0\linewidth]{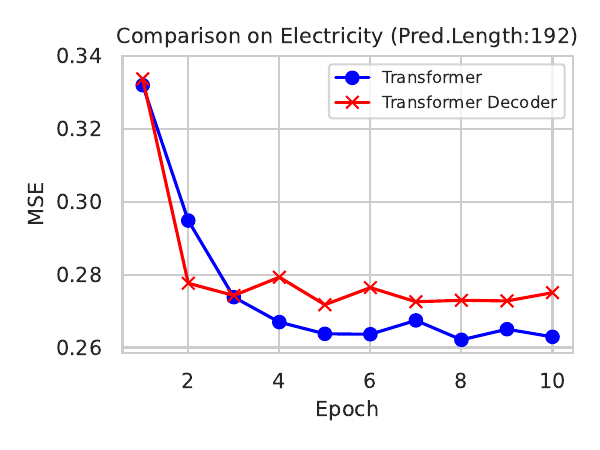}
  \end{subfigure}
  \begin{subfigure}{.5\textwidth}
    \centering
    \includegraphics[width=1.0\linewidth]{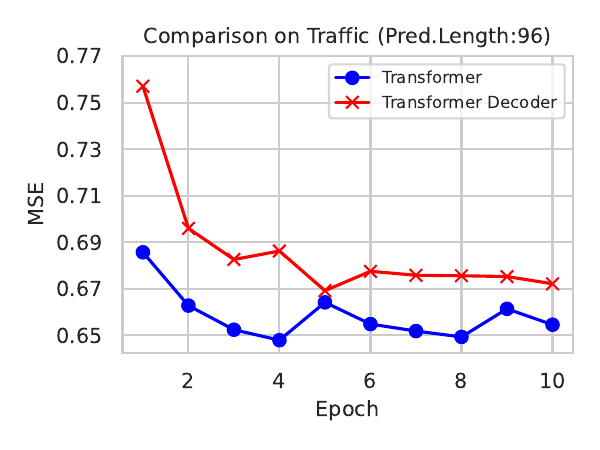}
  \end{subfigure}%
  \begin{subfigure}{.5\textwidth}
      \centering
      \includegraphics[width=1.0\linewidth]{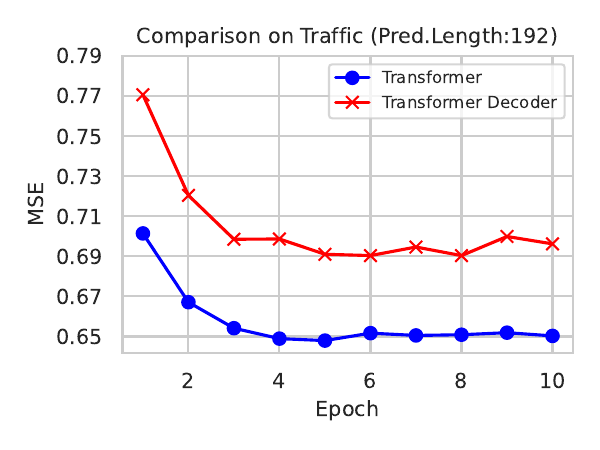}
  \end{subfigure}
  
  \caption{Comparative Analysis of the Original Transformer versus a Decoder-Only Transformer Model on Electricity and Traffic Datasets. The Decoder-Only model operates with a reduced input of merely 8 historical tokens, distinctly fewer than the 48 tokens utilized in the standard Transformer setup. 
  }
  \label{fig:Comp}
\end{figure}

\section{Experiments}
\subsection{Data and experiment setting}
In this study, we evaluate the performance of CVTN using eight popular datasets from various fields, including electricity\cite{misc_electricityloaddiagrams20112014_321}, traffic\cite{pems_traffic}, weather\cite{wetterstation}, four ETT (Electricity Transformer Temperature)\cite{zhou2021informer} and exchange\cite{lai2018modeling}. The look-back window size for all datasets is uniformly set at 96, and the number of training epochs is fixed at 10 for each. We assess the performance using four different prediction lengths (96, 192, 336, and 720). Following the evaluation procedure used in previous studies, we compute the Mean Squared Error (MSE) and Mean Absolute Error (MAE) for data normalized with z-score normalization. This approach allows us to measure different variables on a consistent scale.

\subsection{Transformer Prediction Analysis}
\begin{table}[!htbp]
  \caption{The complete results for LTSF. The results of 4 different prediction lengths of different models are listed in the table. The look-back window sizes are set to 96 for all datasets. We also calculate the average (Avg) and median(Me) of the results for the 4 prediction lengths and the number of optimal values obtained by different models.}\label{tab:com_forecasting_results}
  \vskip 0.05in
  \centering
  \resizebox{1.0\columnwidth}{!}{
  \begin{threeparttable}
  \begin{small}
  \renewcommand{\multirowsetup}{\centering}
  \setlength{\tabcolsep}{1pt}
  \begin{tabular}{c|c|cc|cc|cc|cc|cc|cc|cc|cc|cc|cc|cc}
    \toprule
    \multicolumn{2}{c}{\multirow{2}{*}{Models}} & 
    \multicolumn{2}{c}{\rotatebox{0}{\scalebox{0.8}{\textbf{CVTN}}}} &
    
    \multicolumn{2}{c}{\rotatebox{0}{\scalebox{0.8}{
              \begin{tabular}{@{}c@{}}
                iTransformer \\
              \citeyear{liu2023itransformer}
              \end{tabular}
                }}} &
    \multicolumn{2}{c}{\rotatebox{0}{\scalebox{0.8}{
      \begin{tabular}{@{}c@{}}
        Client \\
      \citeyear{gao2023client}
      \end{tabular}
      }}} &
    \multicolumn{2}{c}{\rotatebox{0}{\scalebox{0.8}{\begin{tabular}{@{}c@{}}
      DLinear \\
      \citeyear{zeng2023transformers}
    \end{tabular}
    }}} & 
    \multicolumn{2}{c}{\rotatebox{0}{\scalebox{0.8}{\begin{tabular}{@{}c@{}}
      TimesNet \\
      \citeyear{wu2022timesnet}
    \end{tabular}
    }}} &
    \multicolumn{2}{c}{\rotatebox{0}{\scalebox{0.8}{
      \begin{tabular}{@{}c@{}}
        FEDformer \\
      \citeyear{zhou2022fedformer}
    \end{tabular}
      }}} & \multicolumn{2}{c}{\rotatebox{0}{\scalebox{0.8}{
        \begin{tabular}{@{}c@{}}
          ETSformer \\
        \citeyear{woo2022etsformer}
        \end{tabular}
        }}} & 
        \multicolumn{2}{c}{\rotatebox{0}{\scalebox{0.8}{
              \begin{tabular}{@{}c@{}}
                LightTS \\
              \citeyear{zhang2022less}
              \end{tabular}
                }}} &
        \multicolumn{2}{c}{\rotatebox{0}{\scalebox{0.8}{
          \begin{tabular}{@{}c@{}}
            Autoformer \\
          \citeyear{wu2021autoformer}
          \end{tabular}
          }}} & 
          \multicolumn{2}{c}{\rotatebox{0}{\scalebox{0.8}{
            \begin{tabular}{@{}c@{}}
              Pyraformer \\
            \citeyear{liu2021pyraformer}
            \end{tabular}
            }}} &  \multicolumn{2}{c}{\rotatebox{0}{\scalebox{0.8}{
              \begin{tabular}{@{}c@{}}
                Informer \\
              \citeyear{zhou2021informer}
              \end{tabular}
              }}} \\
    \cmidrule(lr){3-4} \cmidrule(lr){5-6}\cmidrule(lr){7-8} \cmidrule(lr){9-10}\cmidrule(lr){11-12}\cmidrule(lr){13-14}\cmidrule(lr){15-16}\cmidrule(lr){17-18}\cmidrule(lr){19-20}\cmidrule(lr){21-22}\cmidrule(lr){23-24}

    \multicolumn{2}{c}{Metric} & \scalebox{0.78}{MSE} & \scalebox{0.78}{MAE} & \scalebox{0.78}{MSE} & \scalebox{0.78}{MAE} & \scalebox{0.78}{MSE} & \scalebox{0.78}{MAE} & \scalebox{0.78}{MSE} & \scalebox{0.78}{MAE} & \scalebox{0.78}{MSE} & \scalebox{0.78}{MAE} & \scalebox{0.78}{MSE} & \scalebox{0.78}{MAE} & \scalebox{0.78}{MSE} & \scalebox{0.78}{MAE} & \scalebox{0.78}{MSE} & \scalebox{0.78}{MAE} & \scalebox{0.78}{MSE} & \scalebox{0.78}{MAE} & \scalebox{0.78}{MSE} & \scalebox{0.78}{MAE} & \scalebox{0.78}{MSE} & \scalebox{0.78}{MAE}\\
    \toprule 
    \multirow{5}{*}{\rotatebox{90}{\scalebox{0.95}{Electricity}}} 
    &  \scalebox{0.78}{96} &\boldres{\scalebox{0.78}{0.132}} &\boldres{\scalebox{0.78}{0.230}} &\scalebox{0.78}{0.148} &\scalebox{0.78}{0.240} &\secondres{\scalebox{0.78}{0.141}} &\secondres{\scalebox{0.78}{0.236}} &\scalebox{0.78}{0.197} &\scalebox{0.78}{0.282} &\scalebox{0.78}{0.168} &\scalebox{0.78}{0.272} &\scalebox{0.78}{0.193} &\scalebox{0.78}{0.308} &{\scalebox{0.78}{0.187}} &{\scalebox{0.78}{0.304}} &\scalebox{0.78}{0.207} &\scalebox{0.78}{0.307} &\scalebox{0.78}{0.201} &\scalebox{0.78}{0.317} &\scalebox{0.78}{0.386} &\scalebox{0.78}{0.449} &\scalebox{0.78}{0.274} &\scalebox{0.78}{0.368}  \\

    & \scalebox{0.78}{192} &\boldres{\scalebox{0.78}{0.154}} &\scalebox{0.78}{0.256} &\scalebox{0.78}{0.162} &\boldres{\scalebox{0.78}{0.253}} & \secondres{\scalebox{0.78}{0.161}} &\secondres{\scalebox{0.78}{0.254}} &\scalebox{0.78}{0.196} &\scalebox{0.78}{0.285}&\scalebox{0.78}{0.184} &\scalebox{0.78}{0.289} &\scalebox{0.78}{0.201} &\scalebox{0.78}{0.315} &{\scalebox{0.78}{0.199}} &{\scalebox{0.78}{0.315}} &\scalebox{0.78}{0.213} &\scalebox{0.78}{0.316} &\scalebox{0.78}{0.222} &\scalebox{0.78}{0.334} &\scalebox{0.78}{0.378} &\scalebox{0.78}{0.443} &\scalebox{0.78}{0.296} &\scalebox{0.78}{0.386}\\

    & \scalebox{0.78}{336} &\boldres{\scalebox{0.78}{0.170}} &\scalebox{0.78}{0.274} &\scalebox{0.78}{0.178} &\secondres{\scalebox{0.78}{0.269}} & \secondres{\scalebox{0.78}{0.173}} &\boldres{\scalebox{0.78}{0.267}} &\scalebox{0.78}{0.209} &\scalebox{0.78}{0.301}&\scalebox{0.78}{0.198} &\scalebox{0.78}{0.300} &\scalebox{0.78}{0.214} &\scalebox{0.78}{0.329} &{\scalebox{0.78}{0.212}} &{\scalebox{0.78}{0.329}} &\scalebox{0.78}{0.230} &\scalebox{0.78}{0.333} &\scalebox{0.78}{0.231} &\scalebox{0.78}{0.338} &\scalebox{0.78}{0.376} &\scalebox{0.78}{0.443} &\scalebox{0.78}{0.300} &\scalebox{0.78}{0.394} \\

    & \scalebox{0.78}{720} &\boldres{\scalebox{0.78}{0.187}} &\boldres{\scalebox{0.78}{0.290}} &\scalebox{0.78}{0.225} &\scalebox{0.78}{0.317} & \secondres{{\scalebox{0.78}{0.209}}} &\secondres{\scalebox{0.78}{0.299}} &\scalebox{0.78}{0.245} &\scalebox{0.78}{0.333}&\scalebox{0.78}{0.220} &\scalebox{0.78}{0.320} &\scalebox{0.78}{0.246} &\scalebox{0.78}{0.355} &{{\scalebox{0.78}{0.233}}} &{\scalebox{0.78}{0.245}} &\scalebox{0.78}{0.265} &\scalebox{0.78}{0.360} &\scalebox{0.78}{0.254} &\scalebox{0.78}{0.361} &\scalebox{0.78}{0.376} &\scalebox{0.78}{0.445} &\scalebox{0.78}{0.373} &\scalebox{0.78}{0.439} \\

    \cmidrule(lr){2-24}
    & \scalebox{0.78}{Avg} &\boldres{\scalebox{0.78}{0.161}} &\boldres{\scalebox{0.78}{0.263}} &\scalebox{0.78}{0.178} &\scalebox{0.78}{0.270} &\secondres{\scalebox{0.78}{0.171}}&\secondres{\scalebox{0.78}{0.264}}&\scalebox{0.78}{0.212} &\scalebox{0.78}{0.300} &\scalebox{0.78}{0.192} &\scalebox{0.78}{0.295} &\scalebox{0.78}{0.214} &\scalebox{0.78}{0.327} &{\scalebox{0.78}{0.208}}&{\scalebox{0.78}{0.323}} &\scalebox{0.78}{0.229} &\scalebox{0.78}{0.329} &\scalebox{0.78}{0.227} &\scalebox{0.78}{0.338} &\scalebox{0.78}{0.379} &\scalebox{0.78}{0.445} &\scalebox{0.78}{0.311} &\scalebox{0.78}{0.397} \\

    & \scalebox{0.78}{Me} 
    &\boldres{\scalebox{0.78}{0.162}} 
    &\secondres{\scalebox{0.78}{0.265}} 
    &\scalebox{0.78}{0.170} 
    &\boldres{\scalebox{0.78}{0.261}} 
    &\secondres{\scalebox{0.78}{0.167}}
    &\boldres{\scalebox{0.78}{0.261}} 
    &\scalebox{0.78}{0.203} 
    &\scalebox{0.78}{0.293}
    &\scalebox{0.78}{0.191} 
    &\scalebox{0.78}{0.295} 
    &\scalebox{0.78}{0.208} 
    &\scalebox{0.78}{0.322} 
    &{\scalebox{0.78}{0.206}}
    &{\scalebox{0.78}{0.322}} 
    &\scalebox{0.78}{0.222} 
    &\scalebox{0.78}{0.325} 
    &\scalebox{0.78}{0.227} 
    &\scalebox{0.78}{0.336} 
    &\scalebox{0.78}{0.377} 
    &\scalebox{0.78}{0.444}
    &\scalebox{0.78}{0.298} 
    &\scalebox{0.78}{0.390} \\

    \midrule

    \multirow{5}{*}{\rotatebox{90}{\scalebox{0.95}{Traffic}}} 
    & \scalebox{0.78}{96} &\secondres{\scalebox{0.78}{0.416}} &\boldres{\scalebox{0.78}{0.261}} &\boldres{\scalebox{0.78}{0.395}} &\secondres{\scalebox{0.78}{0.268}} & \scalebox{0.78}{0.438} &\scalebox{0.78}{0.292} &\scalebox{0.78}{0.650} &\scalebox{0.78}{0.396} &\scalebox{0.78}{0.593} &\scalebox{0.78}{0.321} &{\scalebox{0.78}{0.587}} &\scalebox{0.78}{0.366} &\scalebox{0.78}{0.607} &\scalebox{0.78}{0.392} &\scalebox{0.78}{0.615} &\scalebox{0.78}{0.391} &\scalebox{0.78}{0.613} &\scalebox{0.78}{0.388} &\scalebox{0.78}{0.867} &\scalebox{0.78}{0.468} &\scalebox{0.78}{0.719} &\scalebox{0.78}{0.391} \\

    & \scalebox{0.78}{192} &\secondres{\scalebox{0.78}{0.430}} &\boldres{\scalebox{0.78}{0.272}} &\boldres{\scalebox{0.78}{0.417}} &\secondres{\scalebox{0.78}{0.276}} & \scalebox{0.78}{0.451} &\scalebox{0.78}{0.298} &{\scalebox{0.78}{0.598}} &\scalebox{0.78}{0.370} &{\scalebox{0.78}{0.617}} &\scalebox{0.78}{0.336} &\scalebox{0.78}{0.604} &\scalebox{0.78}{0.373} &\scalebox{0.78}{0.621} &\scalebox{0.78}{0.399} &\scalebox{0.78}{0.601} &\scalebox{0.78}{0.382} &\scalebox{0.78}{0.616} &\scalebox{0.78}{0.382} &\scalebox{0.78}{0.869} &\scalebox{0.78}{0.467} &\scalebox{0.78}{0.696} &\scalebox{0.78}{0.379} \\

    & \scalebox{0.78}{336} &\secondres{\scalebox{0.78}{0.444}} &\boldres{\scalebox{0.78}{0.274}} &\boldres{\scalebox{0.78}{0.433}} &\secondres{\scalebox{0.78}{0.283}} & \scalebox{0.78}{0.472} &{\scalebox{0.78}{0.305}} &{\scalebox{0.78}{0.605}} &\scalebox{0.78}{0.373}&{\scalebox{0.78}{0.629}} &\scalebox{0.78}{0.336} &\scalebox{0.78}{0.621} &\scalebox{0.78}{0.383} &\scalebox{0.78}{0.622} &{\scalebox{0.78}{0.399}} &\scalebox{0.78}{0.613} &\scalebox{0.78}{0.386} &\scalebox{0.78}{0.622} &\scalebox{0.78}{0.337} &\scalebox{0.78}{0.881} &\scalebox{0.78}{0.469} &\scalebox{0.78}{0.777} &\scalebox{0.78}{0.420} \\

    & \scalebox{0.78}{720} &\secondres{\scalebox{0.78}{0.482}} &\boldres{\scalebox{0.78}{0.292}} &\boldres{\scalebox{0.78}{0.467}} &\secondres{\scalebox{0.78}{0.302}} & \scalebox{0.78}{0.499} &\scalebox{0.78}{0.321}  &\scalebox{0.78}{0.645} &\scalebox{0.78}{0.394}&\scalebox{0.78}{0.640} &\scalebox{0.78}{0.350} &{\scalebox{0.78}{0.626}} &\scalebox{0.78}{0.382} &\scalebox{0.78}{0.632} &\scalebox{0.78}{0.396} &\scalebox{0.78}{0.658} &\scalebox{0.78}{0.407} &\scalebox{0.78}{0.660} &\scalebox{0.78}{0.408} &\scalebox{0.78}{0.896} &\scalebox{0.78}{0.473} &\scalebox{0.78}{0.864} &\scalebox{0.78}{0.472} \\
    \cmidrule(lr){2-24}
    & \scalebox{0.78}{Avg} &\secondres{\scalebox{0.78}{0.443}} &\boldres{\scalebox{0.78}{0.275}} &\boldres{\scalebox{0.78}{0.428}} &\secondres{\scalebox{0.78}{0.282}} & \scalebox{0.78}{0.465} &\scalebox{0.78}{0.304}  &\scalebox{0.78}{0.625} &\scalebox{0.78}{0.383}&\scalebox{0.78}{0.620} &\scalebox{0.78}{0.336} &{\scalebox{0.78}{0.610}} &\scalebox{0.78}{0.376} &\scalebox{0.78}{0.621} &\scalebox{0.78}{0.396} &\scalebox{0.78}{0.622} &\scalebox{0.78}{0.392} &\scalebox{0.78}{0.628} &\scalebox{0.78}{0.379} &\scalebox{0.78}{0.878} &\scalebox{0.78}{0.469} &\scalebox{0.78}{0.764} &\scalebox{0.78}{0.416} \\
    & \scalebox{0.78}{Me} 
&\secondres{\scalebox{0.78}{0.437}} 
&\boldres{\scalebox{0.78}{0.273}} 
&\boldres{\scalebox{0.78}{0.425}} 
&\secondres{\scalebox{0.78}{0.280}}
&\scalebox{0.78}{0.462}
&\scalebox{0.78}{0.302}
&\scalebox{0.78}{0.625} 
&\scalebox{0.78}{0.384} 
&\scalebox{0.78}{0.623} 
&\scalebox{0.78}{0.336} 
&\scalebox{0.78}{0.613} 
&\scalebox{0.78}{0.378} 
&\scalebox{0.78}{0.622}
&\scalebox{0.78}{0.396}
&\scalebox{0.78}{0.614} 
&\scalebox{0.78}{0.389} 
&\scalebox{0.78}{0.619} 
&\scalebox{0.78}{0.385} 
&\scalebox{0.78}{0.875} 
&\scalebox{0.78}{0.469}
&\scalebox{0.78}{0.748} 
&\scalebox{0.78}{0.406}\\

    \midrule
    \multirow{5}{*}{\rotatebox{90}{\scalebox{0.95}{Weather}}} 
    &  \scalebox{0.78}{96} &\boldres{\scalebox{0.78}{0.153}} &\boldres{\scalebox{0.78}{0.201}} &\scalebox{0.78}{0.174} &\scalebox{0.78}{0.214} &\secondres{\scalebox{0.78}{0.163}} &\secondres{\scalebox{0.78}{0.207}} &\scalebox{0.78}{0.196} &\scalebox{0.78}{0.255} & \scalebox{0.78}{0.172} &\scalebox{0.78}{0.220} & \scalebox{0.78}{0.217} &\scalebox{0.78}{0.296} & \scalebox{0.78}{0.197} &{\scalebox{0.78}{0.281}} & \scalebox{0.78}{0.182} &\scalebox{0.78}{0.242} & \scalebox{0.78}{0.266} &\scalebox{0.78}{0.336} & \scalebox{0.78}{0.622} &\scalebox{0.78}{0.556} & \scalebox{0.78}{0.300} &\scalebox{0.78}{0.384}\\

  & \scalebox{0.78}{192} &\boldres{\scalebox{0.78}{0.204}} &\boldres{\scalebox{0.78}{0.250}} &\scalebox{0.78}{0.221} &\scalebox{0.78}{0.254} & \secondres{\scalebox{0.78}{0.214}} &\secondres{\scalebox{0.78}{0.253}} & \scalebox{0.78}{0.237} &\scalebox{0.78}{0.296} & \scalebox{0.78}{0.219} &\scalebox{0.78}{0.261} & \scalebox{0.78}{0.276} &\scalebox{0.78}{0.336} & {\scalebox{0.78}{0.237}} &\scalebox{0.78}{0.312} & \scalebox{0.78}{0.227} &\scalebox{0.78}{0.287} & \scalebox{0.78}{0.307} &\scalebox{0.78}{0.367} & \scalebox{0.78}{0.739} &\scalebox{0.78}{0.624}  & \scalebox{0.78}{0.598} &\scalebox{0.78}{0.544} \\

  & \scalebox{0.78}{336} &\boldres{\scalebox{0.78}{0.259}} &\boldres{\scalebox{0.78}{0.294}} &\scalebox{0.78}{0.278} &\scalebox{0.78}{0.296} &\secondres{\scalebox{0.78}{0.271}} &\secondres{\scalebox{0.78}{0.294}} &\scalebox{0.78}{0.283} &\scalebox{0.78}{0.335}& \scalebox{0.78}{0.280} &\scalebox{0.78}{0.306} & \scalebox{0.78}{0.339} &\scalebox{0.78}{0.380} & \scalebox{0.78}{0.298} &\scalebox{0.78}{0.353} & \scalebox{0.78}{0.282} &\scalebox{0.78}{0.334} & \scalebox{0.78}{0.359} &\scalebox{0.78}{0.395} & \scalebox{0.78}{1.004} &\scalebox{0.78}{0.753} & \scalebox{0.78}{0.578} &\scalebox{0.78}{0.523} \\
  
& \scalebox{0.78}{720} &\boldres{\scalebox{0.78}{0.331}} &\scalebox{0.78}{0.351} &\scalebox{0.78}{0.358} &\secondres{\scalebox{0.78}{0.349}} & \scalebox{0.78}{0.360} &\boldres{\scalebox{0.78}{0.346}} &\secondres{\scalebox{0.78}{0.345}} &\scalebox{0.78}{0.381} & \scalebox{0.78}{0.365} &\scalebox{0.78}{0.359} & \scalebox{0.78}{0.403} &\scalebox{0.78}{0.428} & \scalebox{0.78}{0.352} &\scalebox{0.78}{0.390} & \scalebox{0.78}{0.352} &\scalebox{0.78}{0.386} & \scalebox{0.78}{0.419} &\scalebox{0.78}{0.428} & \scalebox{0.78}{1.420} &\scalebox{0.78}{0.934} & \scalebox{0.78}{1.059} &\scalebox{0.78}{0.741} \\

\cmidrule(lr){2-24}
& \scalebox{0.78}{Avg} &\boldres{\scalebox{0.78}{0.237}} &\boldres{\scalebox{0.78}{0.274}} &\scalebox{0.78}{0.258} &\scalebox{0.78}{0.279} & \secondres{\scalebox{0.78}{0.249}} &\secondres{\scalebox{0.78}{0.275}}&\scalebox{0.78}{0.265} &\scalebox{0.78}{0.317}&\scalebox{0.78}{0.259} &\scalebox{0.78}{0.287} &\scalebox{0.78}{0.309} &\scalebox{0.78}{0.360} &\scalebox{0.78}{0.271} &\scalebox{0.78}{0.334} &\scalebox{0.78}{0.261} &\scalebox{0.78}{0.312} &\scalebox{0.78}{0.338} &\scalebox{0.78}{0.382} &\scalebox{0.78}{0.946} &\scalebox{0.78}{0.717}&\scalebox{0.78}{0.634} &\scalebox{0.78}{0.548}\\
& \scalebox{0.78}{Me} 
&\boldres{\scalebox{0.78}{0.232}} 
&\boldres{\scalebox{0.78}{0.272}} 
&\scalebox{0.78}{0.250} 
&\scalebox{0.78}{0.275} 
&\secondres{\scalebox{0.78}{0.243}}
&\secondres{\scalebox{0.78}{0.274}} 
&\scalebox{0.78}{0.260} 
&\scalebox{0.78}{0.316} 
&\scalebox{0.78}{0.250} 
&\scalebox{0.78}{0.284} 
&\scalebox{0.78}{0.308} 
&\scalebox{0.78}{0.358} 
&{\scalebox{0.78}{0.268}}
&{\scalebox{0.78}{0.333}} 
&\scalebox{0.78}{0.255} 
&\scalebox{0.78}{0.311} 
&\scalebox{0.78}{0.333} 
&\scalebox{0.78}{0.381} 
&\scalebox{0.78}{0.872} 
&\scalebox{0.78}{0.689} 
&\scalebox{0.78}{0.588} 
&\scalebox{0.78}{0.534} \\

\midrule
\multirow{5}{*}{\rotatebox{90}{\scalebox{0.95}{ETTh1}}}
    &  \scalebox{0.78}{96} &\scalebox{0.78}{0.386} &\boldres{\scalebox{0.78}{0.400}} &\scalebox{0.78}{0.386} &\scalebox{0.78}{0.405} & \scalebox{0.78}{0.392} &\scalebox{0.78}{0.409} &\scalebox{0.78}{0.386} &\boldres{\scalebox{0.78}{0.400}} & \secondres{\scalebox{0.78}{0.384}} &\secondres{\scalebox{0.78}{0.402}} &\boldres{\scalebox{0.78}{0.376}} &\scalebox{0.78}{0.419} &\scalebox{0.78}{0.494} &\scalebox{0.78}{0.479} &\scalebox{0.78}{0.424} &\scalebox{0.78}{0.432} &\scalebox{0.78}{0.449} &\scalebox{0.78}{0.459} &\scalebox{0.78}{0.664} &\scalebox{0.78}{0.612} &\scalebox{0.78}{0.865} &\scalebox{0.78}{0.713}\\

    & \scalebox{0.78}{192} &\scalebox{0.78}{0.443} &\secondres{\scalebox{0.78}{0.431}} &\scalebox{0.78}{0.441} &\scalebox{0.78}{0.436} & \scalebox{0.78}{0.445} &\scalebox{0.78}{0.436} &\scalebox{0.78}{0.437} &\scalebox{0.78}{0.432}&\secondres{\scalebox{0.78}{0.436}} &\boldres{\scalebox{0.78}{0.429}} &\boldres{\scalebox{0.78}{0.420}} &\scalebox{0.78}{0.448} &\scalebox{0.78}{0.538} &\scalebox{0.78}{0.504} &\scalebox{0.78}{0.475} &\scalebox{0.78}{0.462} &\scalebox{0.78}{0.500} &\scalebox{0.78}{0.482} &\scalebox{0.78}{0.790} &\scalebox{0.78}{0.681 }&\scalebox{0.78}{1.008} &\scalebox{0.78}{0.792} \\

    & \scalebox{0.78}{336} &\secondres{\scalebox{0.78}{0.478}} &\boldres{\scalebox{0.78}{0.451}} &\scalebox{0.78}{0.487} &\scalebox{0.78}{0.458}  & \scalebox{0.78}{0.482} &\secondres{\scalebox{0.78}{0.456}} &\scalebox{0.78}{0.481} &\scalebox{0.78}{0.459}&\scalebox{0.78}{0.491} &\scalebox{0.78}{0.469} &\boldres{\scalebox{0.78}{0.459}} &\scalebox{0.78}{0.465} &\scalebox{0.78}{0.574} &{\scalebox{0.78}{0.521}} &\scalebox{0.78}{0.518} &\scalebox{0.78}{0.488} &\scalebox{0.78}{0.521} &\scalebox{0.78}{0.496} &\scalebox{0.78}{0.891} &\scalebox{0.78}{0.738} &\scalebox{0.78}{1.107} &\scalebox{0.78}{0.809}\\

    & \scalebox{0.78}{720} &\boldres{\scalebox{0.78}{0.477}} &\boldres{\scalebox{0.78}{0.467}} &\scalebox{0.78}{0.503} &\scalebox{0.78}{0.491}  & \secondres{\scalebox{0.78}{0.489}} &\secondres{\scalebox{0.78}{0.480}}& \scalebox{0.78}{0.519} &\scalebox{0.78}{0.516}&\scalebox{0.78}{0.521} &\scalebox{0.78}{0.500} &\scalebox{0.78}{0.506} &\scalebox{0.78}{0.507} &{\scalebox{0.78}{0.562}} &{\scalebox{0.78}{0.535}} &\scalebox{0.78}{0.547} &\scalebox{0.78}{0.533} &\scalebox{0.78}{0.514} &\scalebox{0.78}{0.512} &\scalebox{0.78}{0.963} &\scalebox{0.78}{0.782} &\scalebox{0.78}{1.181} &\scalebox{0.78}{0.865}\\
    \cmidrule(lr){2-24}
    & \scalebox{0.78}{Avg} &\secondres{\scalebox{0.78}{0.446}} &\boldres{\scalebox{0.78}{0.437}} &\scalebox{0.78}{0.454} &\scalebox{0.78}{0.447} & \scalebox{0.78}{0.452} &\secondres{\scalebox{0.78}{0.445}}& \scalebox{0.78}{0.456} &\scalebox{0.78}{0.452}& \scalebox{0.78}{0.458} &\scalebox{0.78}{0.450} &\boldres{\scalebox{0.78}{0.440}} &\scalebox{0.78}{0.460} &\scalebox{0.78}{0.542} &{\scalebox{0.78}{0.510}} &\scalebox{0.78}{0.491} &\scalebox{0.78}{0.479} &\scalebox{0.78}{0.496} &\scalebox{0.78}{0.487} &\scalebox{0.78}{0.827} &\scalebox{0.78}{0.703} &\scalebox{0.78}{1.040} &\scalebox{0.78}{0.795}\\
    & \scalebox{0.78}{Me} 
&\scalebox{0.78}{0.460} 
&\boldres{\scalebox{0.78}{0.441}} 
&\scalebox{0.78}{0.464} 
&\scalebox{0.78}{0.447}
&\scalebox{0.78}{0.464}
&\secondres{\scalebox{0.78}{0.446}} 
&\secondres{\scalebox{0.78}{0.459}} 
&\secondres{\scalebox{0.78}{0.446}} 
&{\scalebox{0.78}{0.464}} 
&{\scalebox{0.78}{0.449}} 
&\boldres{\scalebox{0.78}{0.440}} 
&\scalebox{0.78}{0.457} 
&\scalebox{0.78}{0.550}
&{\scalebox{0.78}{0.513}} 
&\scalebox{0.78}{0.497} 
&\scalebox{0.78}{0.475} 
&\scalebox{0.78}{0.507} 
&\scalebox{0.78}{0.489} 
&\scalebox{0.78}{0.841} 
&\scalebox{0.78}{0.710}
&\scalebox{0.78}{1.058} 
&\scalebox{0.78}{0.801} \\

\midrule
\multirow{5}{*}{\rotatebox{90}{\scalebox{0.95}{ETTh2}}}  
&  \scalebox{0.78}{96} & \secondres{\scalebox{0.78}{0.300}} & \secondres{\scalebox{0.78}{0.350}} & \boldres{\scalebox{0.78}{0.297}} &\boldres{\scalebox{0.78}{0.349}} & {\scalebox{0.78}{0.305}} &{\scalebox{0.78}{0.353}} &{\scalebox{0.78}{0.333}} &{\scalebox{0.78}{0.387}}&{\scalebox{0.78}{0.340}} &{\scalebox{0.78}{0.374}}  &\scalebox{0.78}{0.358} &\scalebox{0.78}{0.397} &{\scalebox{0.78}{0.340}} &{\scalebox{0.78}{0.391}} &\scalebox{0.78}{0.397} &\scalebox{0.78}{0.437} &\scalebox{0.78}{0.346} &\scalebox{0.78}{0.388} &\scalebox{0.78}{0.645} &\scalebox{0.78}{0.597} &\scalebox{0.78}{3.755} &\scalebox{0.78}{1.525} \\

& \scalebox{0.78}{192} & \boldres{\scalebox{0.78}{0.369}} & \boldres{\scalebox{0.78}{0.393}} & \secondres{\scalebox{0.78}{0.380}} &\secondres{\scalebox{0.78}{0.400}} &{\scalebox{0.78}{0.382}} &{\scalebox{0.78}{0.401}} &\scalebox{0.78}{0.477} &\scalebox{0.78}{0.476}&\scalebox{0.78}{0.402} &\scalebox{0.78}{0.414} &\scalebox{0.78}{0.429} &\scalebox{0.78}{0.439} &{\scalebox{0.78}{0.430}} &{\scalebox{0.78}{0.439}} &\scalebox{0.78}{0.520} &\scalebox{0.78}{0.504}  &\scalebox{0.78}{0.456} &\scalebox{0.78}{0.452} &\scalebox{0.78}{0.788} &\scalebox{0.78}{0.683} &\scalebox{0.78}{5.602} &\scalebox{0.78}{1.931} \\

& \scalebox{0.78}{336} & \boldres{\scalebox{0.78}{0.419}} & \secondres{\scalebox{0.78}{0.434}} & \secondres{\scalebox{0.78}{0.428}} &\boldres{\scalebox{0.78}{0.432}} &{\scalebox{0.78}{0.434}} &{\scalebox{0.78}{0.445}} &\scalebox{0.78}{0.594} &\scalebox{0.78}{0.541}&\scalebox{0.78}{0.452} &\scalebox{0.78}{0.452} &\scalebox{0.78}{0.496} &\scalebox{0.78}{0.487} &{\scalebox{0.78}{0.485}} &{\scalebox{0.78}{0.479}} &\scalebox{0.78}{0.626} &\scalebox{0.78}{0.559}  &\scalebox{0.78}{0.482} &\scalebox{0.78}{0.486} &\scalebox{0.78}{0.907} &\scalebox{0.78}{0.747} &\scalebox{0.78}{4.721} &\scalebox{0.78}{1.835}\\

& \scalebox{0.78}{720} & \boldres{\scalebox{0.78}{0.420}} & \boldres{\scalebox{0.78}{0.440}} & \scalebox{0.78}{0.427} &\scalebox{0.78}{0.445} & \secondres{\scalebox{0.78}{0.424}} &\secondres{\scalebox{0.78}{0.444}}  &\scalebox{0.78}{0.831} &\scalebox{0.78}{0.657}&\scalebox{0.78}{0.462} &\scalebox{0.78}{0.468} &\scalebox{0.78}{0.463} &\scalebox{0.78}{0.474} &{\scalebox{0.78}{0.500}} &{\scalebox{0.78}{0.497}} &\scalebox{0.78}{0.863} &\scalebox{0.78}{0.672}  &\scalebox{0.78}{0.515} &\scalebox{0.78}{0.511} &\scalebox{0.78}{0.963} &\scalebox{0.78}{0.783} &\scalebox{0.78}{3.647} &\scalebox{0.78}{1.625} \\
\cmidrule(lr){2-24}

& \scalebox{0.78}{Avg} &\boldres{\scalebox{0.78}{0.377}} &\boldres{\scalebox{0.78}{0.404}} &\secondres{\scalebox{0.78}{0.383}} &\secondres{\scalebox{0.78}{0.407}} & {\scalebox{0.78}{0.386}} &{\scalebox{0.78}{0.411}} &\scalebox{0.78}{0.559} &\scalebox{0.78}{0.515}&\scalebox{0.78}{0.414} &\scalebox{0.78}{0.427} &\scalebox{0.78}{0.437} &\scalebox{0.78}{0.449} &{\scalebox{0.78}{0.439}} &{\scalebox{0.78}{0.452}} &\scalebox{0.78}{0.602} &\scalebox{0.78}{0.543} &\scalebox{0.78}{0.450} &\scalebox{0.78}{0.459} &\scalebox{0.78}{0.826} &\scalebox{0.78}{0.703} &\scalebox{0.78}{4.431} &\scalebox{0.78}{1.729} \\
& \scalebox{0.78}{Me} 
&\boldres{\scalebox{0.78}{0.394}} 
&\boldres{\scalebox{0.78}{0.414}} 
&\scalebox{0.78}{0.404} 
&\secondres{\scalebox{0.78}{0.416}}
&\secondres{\scalebox{0.78}{0.403}}
&\scalebox{0.78}{0.423}
&\scalebox{0.78}{0.536} 
&\scalebox{0.78}{0.509}
&\scalebox{0.78}{0.427} 
&\scalebox{0.78}{0.433} 
&\scalebox{0.78}{0.446} 
&\scalebox{0.78}{0.457} 
&{\scalebox{0.78}{0.458}}
&\scalebox{0.78}{0.459}
&\scalebox{0.78}{0.573} 
&\scalebox{0.78}{0.532} 
&\scalebox{0.78}{0.469} 
&\scalebox{0.78}{0.469} 
&\scalebox{0.78}{0.848} 
&\scalebox{0.78}{0.715} 
&\scalebox{0.78}{4.238} 
&\scalebox{0.78}{1.730} \\

\midrule

\multirow{5}{*}{\rotatebox{90}{\scalebox{0.95}{ETTm1}}}
&  \scalebox{0.78}{96} &\boldres{\scalebox{0.78}{0.324}} &\boldres{\scalebox{0.78}{0.355}} &\secondres{\scalebox{0.78}{0.334}} &\secondres{\scalebox{0.78}{0.368}} & {\scalebox{0.78}{0.336}} &{\scalebox{0.78}{0.369}} &\scalebox{0.78}{0.345} &{\scalebox{0.78}{0.372}}&\scalebox{0.78}{0.338} &{\scalebox{0.78}{0.375}}  &\scalebox{0.78}{0.379} &\scalebox{0.78}{0.419} &{\scalebox{0.78}{0.375}} &{\scalebox{0.78}{0.398}} &\scalebox{0.78}{0.374} &\scalebox{0.78}{0.409} &\scalebox{0.78}{0.505} &\scalebox{0.78}{0.475} &\scalebox{0.78}{0.543} &\scalebox{0.78}{0.510}&\scalebox{0.78}{0.672} &\scalebox{0.78}{0.571} \\

& \scalebox{0.78}{192} &\boldres{\scalebox{0.78}{0.368}} &\boldres{\scalebox{0.78}{0.379}}  &\scalebox{0.78}{0.377} &\scalebox{0.78}{0.391} &\secondres{\scalebox{0.78}{0.374}} &\secondres{\scalebox{0.78}{0.387}} &{\scalebox{0.78}{0.380}} &{\scalebox{0.78}{0.389}} &{\scalebox{0.78}{0.374}} &{\scalebox{0.78}{0.387}} &\scalebox{0.78}{0.426} &\scalebox{0.78}{0.441} &{\scalebox{0.78}{0.408}} &{\scalebox{0.78}{0.410}} &\scalebox{0.78}{0.400} &\scalebox{0.78}{0.407} &\scalebox{0.78}{0.553} &\scalebox{0.78}{0.496} &\scalebox{0.78}{0.557} &\scalebox{0.78}{0.537}&\scalebox{0.78}{0.795} &\scalebox{0.78}{0.669} \\

& \scalebox{0.78}{336} &\boldres{\scalebox{0.78}{0.397}} &\boldres{\scalebox{0.78}{0.399}} &\scalebox{0.78}{0.426} &\scalebox{0.78}{0.420}  &\secondres{\scalebox{0.78}{0.408}} &\secondres{\scalebox{0.78}{0.407}}&\scalebox{0.78}{0.413} &\scalebox{0.78}{0.413}&\scalebox{0.78}{0.410} &\scalebox{0.78}{0.411} &\scalebox{0.78}{0.445} &\scalebox{0.78}{0.459} &{\scalebox{0.78}{0.435}} &{\scalebox{0.78}{0.428}} &\scalebox{0.78}{0.438} &\scalebox{0.78}{0.438} &\scalebox{0.78}{0.621} &\scalebox{0.78}{0.537}&\scalebox{0.78}{0.754} &\scalebox{0.78}{0.655} &\scalebox{0.78}{1.212} &\scalebox{0.78}{0.871}\\

& \scalebox{0.78}{720} &\secondres{\scalebox{0.78}{0.476}} &\secondres{\scalebox{0.78}{0.443}} &\scalebox{0.78}{0.491} &\scalebox{0.78}{0.459}  &{\scalebox{0.78}{0.477}} &\boldres{\scalebox{0.78}{0.442}} &\boldres{{\scalebox{0.78}{0.474}}} &\scalebox{0.78}{0.453} &{\scalebox{0.78}{0.478}} &\scalebox{0.78}{0.450} &\scalebox{0.78}{0.543} &\scalebox{0.78}{0.490} &{\scalebox{0.78}{0.499}} &{\scalebox{0.78}{0.462}} &\scalebox{0.78}{0.527} &\scalebox{0.78}{0.502} &\scalebox{0.78}{0.671} &\scalebox{0.78}{0.561}&\scalebox{0.78}{0.908} &\scalebox{0.78}{0.724} &\scalebox{0.78}{1.166} &\scalebox{0.78}{0.823}\\
\cmidrule(lr){2-24}
& \scalebox{0.78}{Avg} &\boldres{\scalebox{0.78}{0.391}} &\boldres{\scalebox{0.78}{0.394}} &\scalebox{0.78}{0.407} &\scalebox{0.78}{0.410}  &\secondres{{\scalebox{0.78}{0.399}}} &\secondres{{\scalebox{0.78}{0.401}}}&\scalebox{0.78}{0.403} &\scalebox{0.78}{0.407}&\scalebox{0.78}{0.400} &\scalebox{0.78}{0.406} &\scalebox{0.78}{0.448} &\scalebox{0.78}{0.452} &{{\scalebox{0.78}{0.429}}} &{{\scalebox{0.78}{0.425}}} &\scalebox{0.78}{0.435} &\scalebox{0.78}{0.437} &\scalebox{0.78}{0.588} &\scalebox{0.78}{0.517}&\scalebox{0.78}{0.691} &\scalebox{0.78}{0.607} &\scalebox{0.78}{0.961} &\scalebox{0.78}{0.734}\\
& \scalebox{0.78}{Me} 
&\boldres{\scalebox{0.78}{0.383}} 
&\boldres{\scalebox{0.78}{0.389}} 
&\scalebox{0.78}{0.402} 
&\scalebox{0.78}{0.406} 
&\secondres{\scalebox{0.78}{0.391}}
&\secondres{\scalebox{0.78}{0.397}} 
&\scalebox{0.78}{0.397} 
&\scalebox{0.78}{0.401} 
&\scalebox{0.78}{0.392} 
&\scalebox{0.78}{0.399} 
&\scalebox{0.78}{0.436} 
&\scalebox{0.78}{0.450} 
&{\scalebox{0.78}{0.422}}
&{\scalebox{0.78}{0.419}} 
&\scalebox{0.78}{0.419} 
&\scalebox{0.78}{0.423} 
&\scalebox{0.78}{0.587} 
&\scalebox{0.78}{0.517} 
&\scalebox{0.78}{0.656} 
&\scalebox{0.78}{0.596} 
&\scalebox{0.78}{0.981} 
&\scalebox{0.78}{0.746}  \\
\midrule

\multirow{5}{*}{\rotatebox{90}{\scalebox{0.95}{ETTm2}}}
&  \scalebox{0.78}{96} &\scalebox{0.78}{0.185} &\boldres{\scalebox{0.78}{0.262}} &\boldres{\scalebox{0.78}{0.180}} &\secondres{\scalebox{0.78}{0.264}} &\secondres{\scalebox{0.78}{0.184}} &{\scalebox{0.78}{0.267}}& \scalebox{0.78}{0.193} &\scalebox{0.78}{0.292}  &\scalebox{0.78}{0.187} &\scalebox{0.78}{0.267} &\scalebox{0.78}{0.203} &\scalebox{0.78}{0.287} &{\scalebox{0.78}{0.189}} &{\scalebox{0.78}{0.280}} &\scalebox{0.78}{0.209} &\scalebox{0.78}{0.308} &\scalebox{0.78}{0.255} &\scalebox{0.78}{0.339} &\scalebox{0.78}{0.435} &\scalebox{0.78}{0.507}&\scalebox{0.78}{0.365} &\scalebox{0.78}{0.453} \\

& \scalebox{0.78}{192} &\boldres{\scalebox{0.78}{0.246}} &\boldres{\scalebox{0.78}{0.306}} &\scalebox{0.78}{0.250} &\scalebox{0.78}{0.309} & {\scalebox{0.78}{0.252}} &\secondres{\scalebox{0.78}{0.307}} & \scalebox{0.78}{0.284} &\scalebox{0.78}{0.362} &\secondres{\scalebox{0.78}{0.249}} &\scalebox{0.78}{0.309} &\scalebox{0.78}{0.269} &\scalebox{0.78}{0.328} &{\scalebox{0.78}{0.253}} &{\scalebox{0.78}{0.319}} &\scalebox{0.78}{0.311} &\scalebox{0.78}{0.382} &\scalebox{0.78}{0.281} &\scalebox{0.78}{0.340}&\scalebox{0.78}{0.730} &\scalebox{0.78}{0.673} &\scalebox{0.78}{0.533} &\scalebox{0.78}{0.563} \\

& \scalebox{0.78}{336} &\boldres{\scalebox{0.78}{0.307}} &\boldres{\scalebox{0.78}{0.340}} &\secondres{\scalebox{0.78}{0.311}} &\scalebox{0.78}{0.348}  & {\scalebox{0.78}{0.314}} &\secondres{\scalebox{0.78}{0.345}} &\scalebox{0.78}{0.369} &\scalebox{0.78}{0.427} &\scalebox{0.78}{0.321} &\scalebox{0.78}{0.351} &\scalebox{0.78}{0.325} &\scalebox{0.78}{0.366} &{\scalebox{0.78}{0.314}} &{\scalebox{0.78}{0.357}} &\scalebox{0.78}{0.442} &\scalebox{0.78}{0.446} &\scalebox{0.78}{0.339} &\scalebox{0.78}{0.372} &\scalebox{0.78}{1.201} &\scalebox{0.78}{0.845} &\scalebox{0.78}{1.363} &\scalebox{0.78}{0.887}\\

& \scalebox{0.78}{720} &\boldres{\scalebox{0.78}{0.408}} &\secondres{\scalebox{0.78}{0.403}} &\secondres{\scalebox{0.78}{0.412}} &\scalebox{0.78}{0.407} &\secondres{\scalebox{0.78}{0.412}} &\boldres{\scalebox{0.78}{0.402}}& \scalebox{0.78}{0.554} &\scalebox{0.78}{0.522} &\boldres{\scalebox{0.78}{0.408}} &\secondres{\scalebox{0.78}{0.403}} &\scalebox{0.78}{0.421} &\scalebox{0.78}{0.415} &{\scalebox{0.78}{0.414}} &{\scalebox{0.78}{0.413}} &\scalebox{0.78}{0.675} &\scalebox{0.78}{0.587} &\scalebox{0.78}{0.433} &\scalebox{0.78}{0.432} &\scalebox{0.78}{3.625} &\scalebox{0.78}{1.451} &\scalebox{0.78}{3.379} &\scalebox{0.78}{1.338} \\
\cmidrule(lr){2-24}
& \scalebox{0.78}{Avg} &\boldres{\scalebox{0.78}{0.286}} &\boldres{\scalebox{0.78}{0.328}} &\secondres{\scalebox{0.78}{0.288}} &\scalebox{0.78}{0.332} & {\scalebox{0.78}{0.291}} &\secondres{\scalebox{0.78}{0.330}} & \scalebox{0.78}{0.350} &\scalebox{0.78}{0.401} &\scalebox{0.78}{0.291} &\scalebox{0.78}{0.333} &\scalebox{0.78}{0.305} &\scalebox{0.78}{0.349} &{\scalebox{0.78}{0.293}} &{\scalebox{0.78}{0.342}} &\scalebox{0.78}{0.409} &\scalebox{0.78}{0.436} &\scalebox{0.78}{0.327} &\scalebox{0.78}{0.371} &\scalebox{0.78}{1.498} &\scalebox{0.78}{0.869} &\scalebox{0.78}{1.410} &\scalebox{0.78}{0.810} \\
& \scalebox{0.78}{Me} 
&\boldres{\scalebox{0.78}{0.277}} 
&\boldres{\scalebox{0.78}{0.323}} 
&\secondres{\scalebox{0.78}{0.281}} 
&\scalebox{0.78}{0.329}
&\scalebox{0.78}{0.283}
&\secondres{\scalebox{0.78}{0.326}} 
&\scalebox{0.78}{0.327} 
&\scalebox{0.78}{0.395}
&\scalebox{0.78}{0.285} 
&\scalebox{0.78}{0.330} 
&\scalebox{0.78}{0.297} 
&\scalebox{0.78}{0.347} 
&\scalebox{0.78}{0.284}
&{\scalebox{0.78}{0.338}} 
&\scalebox{0.78}{0.377} 
&\scalebox{0.78}{0.424} 
&\scalebox{0.78}{0.310} 
&\scalebox{0.78}{0.356} 
&\scalebox{0.78}{0.966} 
&\scalebox{0.78}{0.759} 
&\scalebox{0.78}{0.948} 
&\scalebox{0.78}{0.725} \\

\midrule
\multirow{5}{*}{\rotatebox{90}{\scalebox{0.95}{Exchange}}} 
&  \scalebox{0.78}{96} &\boldres{\scalebox{0.78}{0.084}} &\scalebox{0.78}{0.208} &\scalebox{0.78}{0.086} &\secondres{\scalebox{0.78}{0.206}} & {\scalebox{0.78}{0.086}} &\secondres{\scalebox{0.78}{0.206}}  &\scalebox{0.78}{0.088} &\scalebox{0.78}{0.218}&\scalebox{0.78}{0.107} &\scalebox{0.78}{0.234} &\scalebox{0.78}{0.148} &\scalebox{0.78}{0.278} &\secondres{\scalebox{0.78}{0.085}} &\boldres{\scalebox{0.78}{0.204}} &\scalebox{0.78}{0.116} &\scalebox{0.78}{0.262} &\scalebox{0.78}{0.197} &\scalebox{0.78}{0.323} &\scalebox{0.78}{1.748} &\scalebox{0.78}{1.105} &\scalebox{0.78}{0.847} &\scalebox{0.78}{0.752} \\

& \scalebox{0.78}{192}  &\scalebox{0.78}{0.190} &\scalebox{0.78}{0.315} &\secondres{\scalebox{0.78}{0.177}} &\boldres{\scalebox{0.78}{0.299}} & \boldres{\scalebox{0.78}{0.176}} &\boldres{\scalebox{0.78}{0.299}} &\boldres{\scalebox{0.78}{0.176}} &\scalebox{0.78}{0.315}&{\scalebox{0.78}{0.226}} &\scalebox{0.78}{0.334} &\scalebox{0.78}{0.271} &\scalebox{0.78}{0.380} &{\scalebox{0.78}{0.182}} &\secondres{\scalebox{0.78}{0.303}} &\scalebox{0.78}{0.215} &\scalebox{0.78}{0.359} &\scalebox{0.78}{0.300} &\scalebox{0.78}{0.369} &\scalebox{0.78}{1.874} &\scalebox{0.78}{1.151} &\scalebox{0.78}{1.204} &\scalebox{0.78}{0.895} \\

& \scalebox{0.78}{336}  &\scalebox{0.78}{0.345} & \scalebox{0.78}{0.434}&\scalebox{0.78}{0.331} &\secondres{\scalebox{0.78}{0.417}} &\secondres{\scalebox{0.78}{0.330}} & \boldres{\scalebox{0.78}{0.416}}  &\boldres{\scalebox{0.78}{0.313}} &{\scalebox{0.78}{0.427}}&{\scalebox{0.78}{0.367}} &{\scalebox{0.78}{0.448}} &\scalebox{0.78}{0.460} &\scalebox{0.78}{0.500} &{\scalebox{0.78}{0.348}} & {\scalebox{0.78}{0.428}} &\scalebox{0.78}{0.377} &\scalebox{0.78}{0.466} &\scalebox{0.78}{0.509} &\scalebox{0.78}{0.524} &\scalebox{0.78}{1.943} &\scalebox{0.78}{1.172} &\scalebox{0.78}{1.672} &\scalebox{0.78}{1.036} \\

& \scalebox{0.78}{720}  &\boldres{\scalebox{0.78}{0.803}} &\boldres{\scalebox{0.78}{0.679}} &\scalebox{0.78}{0.847} &\scalebox{0.78}{0.691} & \secondres{\scalebox{0.78}{0.828}} &\secondres{\scalebox{0.78}{0.689}} &\scalebox{0.78}{0.839} &{\scalebox{0.78}{0.695}} &\scalebox{0.78}{0.964} &{\scalebox{0.78}{0.746}} &\scalebox{0.78}{1.195} &\scalebox{0.78}{0.841} &{\scalebox{0.78}{1.025}} &{\scalebox{0.78}{0.774}} &\scalebox{0.78}{0.831} &\scalebox{0.78}{0.699} &\scalebox{0.78}{1.447} &\scalebox{0.78}{0.941} &\scalebox{0.78}{2.085} &\scalebox{0.78}{1.206} &\scalebox{0.78}{2.478} &\scalebox{0.78}{1.310}\\
\cmidrule(lr){2-24}
& \scalebox{0.78}{Avg} &\secondres{\scalebox{0.78}{0.355}} &\secondres{\scalebox{0.78}{0.409}} &\scalebox{0.78}{0.360} &\boldres{\scalebox{0.78}{0.403}} & \secondres{\scalebox{0.78}{0.355}} &\boldres{\scalebox{0.78}{0.403}}  &\boldres{\scalebox{0.78}{0.354}} &{\scalebox{0.78}{0.414}} &{\scalebox{0.78}{0.416}} &{\scalebox{0.78}{0.443}} &\scalebox{0.78}{0.519} &\scalebox{0.78}{0.500} &{\scalebox{0.78}{0.410}} &{\scalebox{0.78}{0.427}} &\scalebox{0.78}{0.385} &\scalebox{0.78}{0.447} &\scalebox{0.78}{0.613} &\scalebox{0.78}{0.539} &\scalebox{0.78}{1.913} &\scalebox{0.78}{1.159}  &\scalebox{0.78}{1.550} &\scalebox{0.78}{0.998} \\
& \scalebox{0.78}{Me} 
&\scalebox{0.78}{0.268}
&\scalebox{0.78}{0.375}
&\scalebox{0.78}{0.254} 
&\boldres{\scalebox{0.78}{0.358}}
&\secondres{\scalebox{0.78}{0.253}}
&\boldres{\scalebox{0.78}{0.358}} 
&\boldres{\scalebox{0.78}{0.245}} 
&\scalebox{0.78}{0.371} 
&\scalebox{0.78}{0.297} 
&\scalebox{0.78}{0.396} 
&\scalebox{0.78}{0.366} 
&\scalebox{0.78}{0.440} 
&{\scalebox{0.78}{0.265}}
&\secondres{\scalebox{0.78}{0.366}} 
&\scalebox{0.78}{0.296} 
&\scalebox{0.78}{0.413} 
&\scalebox{0.78}{0.405} 
&\scalebox{0.78}{0.447}
&\scalebox{0.78}{1.909} 
&\scalebox{0.78}{1.162}
&\scalebox{0.78}{1.438} 
&\scalebox{0.78}{0.966}    \\

\midrule

\multicolumn{2}{c}{\scalebox{0.78}{{$1^{\text{st}}$ Count}}} & \multicolumn{2}{c}{\boldres{\scalebox{0.78}{64}}} & \multicolumn{2}{c}{\secondres{\scalebox{0.78}{15}}} & \multicolumn{2}{c}{\scalebox{0.78}{10}} & \multicolumn{2}{c}{\scalebox{0.78}{6}} & \multicolumn{2}{c}{\scalebox{0.78}{2}} & \multicolumn{2}{c}{\scalebox{0.78}{5}} & \multicolumn{2}{c}{\scalebox{0.78}{1}} &  \multicolumn{2}{c}{\scalebox{0.78}{0}} &  \multicolumn{2}{c}{\scalebox{0.78}{0}} & \multicolumn{2}{c}{\scalebox{0.78}{0}} & \multicolumn{2}{c}{{\scalebox{0.78}{0}}} \\

\multicolumn{2}{c}{\scalebox{0.78}{{$2^{\text{st}}$ Count}}} & \multicolumn{2}{c}{{\scalebox{0.78}{18}}} & \multicolumn{2}{c}{\secondres{\scalebox{0.78}{24}}} & \multicolumn{2}{c}{\boldres{\scalebox{0.78}{48}}} & \multicolumn{2}{c}{\scalebox{0.78}{3}} & \multicolumn{2}{c}{\scalebox{0.78}{5}} & \multicolumn{2}{c}{\scalebox{0.78}{0}} & \multicolumn{2}{c}{\scalebox{0.78}{3}} &  \multicolumn{2}{c}{\scalebox{0.78}{0}} &  \multicolumn{2}{c}{\scalebox{0.78}{0}} & \multicolumn{2}{c}{\scalebox{0.78}{0}} & \multicolumn{2}{c}{{\scalebox{0.78}{0}}} \\

\multicolumn{2}{c}{\scalebox{0.78}{{Avg $1^{\text{st}}$ Count}}} & \multicolumn{2}{c}{\boldres{\scalebox{0.78}{12}}} & \multicolumn{2}{c}{\secondres{\scalebox{0.78}{2}}} & \multicolumn{2}{c}{{\scalebox{0.78}{1}}} & \multicolumn{2}{c}{\scalebox{0.78}{1}} & \multicolumn{2}{c}{\scalebox{0.78}{0}} & \multicolumn{2}{c}{\scalebox{0.78}{0}} & \multicolumn{2}{c}{\scalebox{0.78}{0}} &  \multicolumn{2}{c}{\scalebox{0.78}{0}} &  \multicolumn{2}{c}{\scalebox{0.78}{0}} & \multicolumn{2}{c}{\scalebox{0.78}{0}} & \multicolumn{2}{c}{{\scalebox{0.78}{0}}} \\

\multicolumn{2}{c}{\scalebox{0.78}{{Me $1^{\text{st}}$ Count}}} & \multicolumn{2}{c}{\boldres{\scalebox{0.78}{11}}} & \multicolumn{2}{c}{\secondres{\scalebox{0.78}{3}}} & \multicolumn{2}{c}{{\scalebox{0.78}{2}}} & \multicolumn{2}{c}{\scalebox{0.78}{1}} & \multicolumn{2}{c}{\scalebox{0.78}{0}} & \multicolumn{2}{c}{\scalebox{0.78}{0}} & \multicolumn{2}{c}{\scalebox{0.78}{0}} &  \multicolumn{2}{c}{\scalebox{0.78}{0}} &  \multicolumn{2}{c}{\scalebox{0.78}{0}} & \multicolumn{2}{c}{\scalebox{0.78}{0}} & \multicolumn{2}{c}{{\scalebox{0.78}{0}}} \\

    \bottomrule
  \end{tabular}
    \end{small}
  \end{threeparttable}
    }  
\end{table}

In the context of time series prediction problems based on Transformer models, we can perceive the data-driven learning of the Transformer model as two distinct parts. The first part involves the encoder extracting valuable information from historical sequences through self-attention and FFN. The second part is the decoder, which, in conjunction with the encoder's output, models the associative relationships of the target sequence. 

To investigate which part primarily contributes to the Transformer model's benefits, we conducted an extreme experiment. This study tested the original Transformer model and a model using only the Transformer decoder on the Electricity and Traffic datasets. For the decoder-only model, we retained only 8 historical sequences as start tokens for the Transformer's decoder (compared to the original Transformer model, which uses a history sequence length of 48 as start tokens), thereby minimizing the use of historical sequence information as much as possible.

As show in Figure\ref{fig:TSFT} When applying the original Transformer model to time series prediction, we observed significant overfitting. As shown in the figure, despite setting a relatively small learning rate (\(1 \times 10^{-4}\)
), it's apparent that there's an early occurrence of the training set loss decreasing while the validation set loss increases. Moreover, the losses for both the validation and test sets stabilize quickly.

Surprisingly, the model's performance, as depicted in Figure\ref{fig:Comp}, demonstrates that even with a significant reduction in historical information, it achieves a MSE comparable to that of the original Transformer model. This suggests that the original Transformer model did not effectively mine useful information from historical data and indicates that the primary benefit of the Transformer model lies in modeling the associative relationships of the target sequence. Furthermore, previous research\cite{gao2023client} found that obscuring 50\% of the historical input sequence did not significantly degrade the performance of Transformer-based prediction models, further validating this viewpoint.

\subsection{CVTN Prediction Results}
We compared the latest TSFT\cite{wu2022timesnet,gao2023client, zhang2022less,zhou2022fedformer,liu2021pyraformer, Zhou_Zhang_Peng_Zhang_Li_Xiong_Zhang_2022,liu2023itransformer, wu2021autoformer}, CNN-based\cite{wu2022timesnet}, and linear models\cite{zeng2023transformers}. The long-term sequence forecasting results are presented in Table \ref{tab:com_forecasting_results}. The results of the baseline models are referenced from the corresponding papers\cite{wu2022timesnet,gao2023client,liu2023itransformer}.We maintained consistency in the look-back window (96) and training epochs (10) to ensure the most equitable comparison. 

Both iTransformer and Client adopt a cross-variable Transformer architecture, and their performance is only second to CVTN. This indicates the effectiveness of cross-variable learning, as well as the supplementary effectiveness of cross-time learning. DLinear performs well on the Exchange dataset, which has fewer variables, suggesting that DLinear might be effective in certain time series forecasting scenarios that depend on single variables. FEDformer employs frequency domain analysis to process time series data and has shown certain advantages on the ETTh1 dataset, implying that frequency domain features may play a significant role. TimesNet converts time series into multiple two-dimensional tensors to capture their intra-periodic changes and inter-periodic variations, which aligns with our advocacy of focusing on both the periodicity and locality of the sequences. This enables it to perform well on ETTh1 and ETTm2.

From the experimental results, CVTN surpasses many state-of-the-art models, achieving the best performance on several popular datasets. Overall, it achieved first place in 64 categories, and it leads other advanced models by a significant margin in both the average and median numbers of first places in MSE and MAE.

We believe the effectiveness of CVTN stems from several aspects. Firstly, based on our experiments, we identified that the bottleneck of the traditional Transformer model lies in the ineffective utilization of historical sequence information, and its main benefit is in learning the temporal dependency patterns of the prediction sequence. CVTN, through its CVE, thoroughly mines effective information from historical sequences, and through its CTE, fully learns the temporal dependencies of the prediction sequence. Additionally, by separating and training cross-variable and cross-time learning, we avoided the mixing of the two learning modes, which enhanced the prediction results.

\section{Conclusion}
Time series prediction can be decomposed into two types of learning: learning from historical sequences and learning from prediction sequences. For these two learning modes, the traditional Transformer architecture faces two main issues: ineffective learning from historical sequences and overfitting in learning prediction sequences. Most of its benefits come from learning the temporal dependencies in prediction sequences. To achieve effective learning from historical sequences while retaining the benefits of learning temporal dependencies in prediction sequences, this paper proposes CVTN, which divides time series prediction into two stages: cross-variable learning and cross-temporal learning. Cross-variable learning effectively mines information from historical sequences, while cross-temporal learning retains the benefits of learning temporal dependencies in prediction sequences. Moreover, the separation of cross-variable learning and cross-temporal learning avoids the interference of overfitting in cross-temporal learning on cross-variable learning. Experimentally, CVTN achieves state-of-the-art performance on various real-world datasets, which proves the effectiveness of this two-stage algorithm. Our proposed model and findings offer insightful value for the task of time series forecasting.

\section*{Acknowledgement}
This work was supported by the National Natural Science Foundation of China (Grant No. 62106116), China Meteorological Administration Climate Change Special Program (CMA-CCSP) under Grant QBZ202316 and Natural Science
Foundation of Ningbo of China (No. 2023J027)

\small
\bibliographystyle{plain}
\bibliography{neurips_2023}

\begin{thebibliography}{10}

\bibitem{pems_traffic}
Pems: Traffic.
\newblock \url{http://pems.dot.ca.gov/}.

\bibitem{gao2023adaptive}
Jiaxin Gao, Yuntian Chen, Wenbo Hu, and Dongxiao Zhang.
\newblock An adaptive deep-learning load forecasting framework by integrating
  transformer and domain knowledge.
\newblock {\em Advances in Applied Energy}, 10:100142, 2023.

\bibitem{gao2023client}
Jiaxin Gao, Wenbo Hu, and Yuntian Chen.
\newblock Client: Cross-variable linear integrated enhanced transformer for
  multivariate long-term time series forecasting.
\newblock {\em arXiv preprint arXiv:2305.18838}, 2023.

\bibitem{kim2021reversible}
Taesung Kim, Jinhee Kim, Yunwon Tae, Cheonbok Park, Jang-Ho Choi, and Jaegul
  Choo.
\newblock Reversible instance normalization for accurate time-series
  forecasting against distribution shift.
\newblock In {\em International Conference on Learning Representations}, 2021.

\bibitem{lai2018modeling}
Guokun Lai, Wei-Cheng Chang, Yiming Yang, and Hanxiao Liu.
\newblock Modeling long-and short-term temporal patterns with deep neural
  networks.
\newblock In {\em The 41st international ACM SIGIR conference on research \&
  development in information retrieval}, pages 95--104, 2018.

\bibitem{li2019enhancing}
Shiyang Li, Xiaoyong Jin, Yao Xuan, Xiyou Zhou, Wenhu Chen, Yu-Xiang Wang, and
  Xifeng Yan.
\newblock Enhancing the locality and breaking the memory bottleneck of
  transformer on time series forecasting.
\newblock {\em Advances in neural information processing systems}, 32, 2019.

\bibitem{liu2021pyraformer}
Shizhan Liu, Hang Yu, Cong Liao, Jianguo Li, Weiyao Lin, Alex~X Liu, and
  Schahram Dustdar.
\newblock Pyraformer: Low-complexity pyramidal attention for long-range time
  series modeling and forecasting.
\newblock In {\em International conference on learning representations}, 2021.

\bibitem{liu2023itransformer}
Yong Liu, Tengge Hu, Haoran Zhang, Haixu Wu, Shiyu Wang, Lintao Ma, and
  Mingsheng Long.
\newblock itransformer: Inverted transformers are effective for time series
  forecasting.
\newblock {\em arXiv preprint arXiv:2310.06625}, 2023.

\bibitem{liu2022non}
Yong Liu, Haixu Wu, Jianmin Wang, and Mingsheng Long.
\newblock Non-stationary transformers: Exploring the stationarity in time
  series forecasting.
\newblock {\em Advances in Neural Information Processing Systems},
  35:9881--9893, 2022.

\bibitem{lopez2023can}
Alejandro Lopez-Lira and Yuehua Tang.
\newblock Can chatgpt forecast stock price movements? return predictability and
  large language models.
\newblock {\em arXiv preprint arXiv:2304.07619}, 2023.

\bibitem{wetterstation}
{Max-Planck-Institut für Biogeochemie}.
\newblock Wetterstation beutenberg campus.
\newblock \url{https://www.bgc-jena.mpg.de/wetter/}.
\newblock Accessed: [your access date].

\bibitem{meenal2022weather}
R~Meenal, D~Binu, KC~Ramya, Prawin~Angel Michael, K~Vinoth~Kumar,
  E~Rajasekaran, and B~Sangeetha.
\newblock Weather forecasting for renewable energy system: A review.
\newblock {\em Archives of Computational Methods in Engineering},
  29(5):2875--2891, 2022.

\bibitem{mejia2020prediction}
Jose Mejia, Alberto Ochoa-Zezzatti, Oliverio Cruz-Mej{\'\i}a, and Boris
  Mederos.
\newblock Prediction of time series using wavelet gaussian process for wireless
  sensor networks.
\newblock {\em Wireless Networks}, 26(8):5751--5758, 2020.

\bibitem{misc_electricityloaddiagrams20112014_321}
Artur Trindade.
\newblock {ElectricityLoadDiagrams20112014}.
\newblock UCI Machine Learning Repository, 2015.
\newblock {DOI}: https://doi.org/10.24432/C58C86.

\bibitem{vaswani2017attention}
Ashish Vaswani, Noam Shazeer, Niki Parmar, Jakob Uszkoreit, Llion Jones,
  Aidan~N Gomez, {\L}ukasz Kaiser, and Illia Polosukhin.
\newblock Attention is all you need.
\newblock {\em Advances in neural information processing systems}, 30, 2017.

\bibitem{woo2022etsformer}
Gerald Woo, Chenghao Liu, Doyen Sahoo, Akshat Kumar, and Steven Hoi.
\newblock Etsformer: Exponential smoothing transformers for time-series
  forecasting.
\newblock {\em arXiv preprint arXiv:2202.01381}, 2022.

\bibitem{wu2022timesnet}
Haixu Wu, Tengge Hu, Yong Liu, Hang Zhou, Jianmin Wang, and Mingsheng Long.
\newblock Timesnet: Temporal 2d-variation modeling for general time series
  analysis.
\newblock {\em arXiv preprint arXiv:2210.02186}, 2022.

\bibitem{wu2021autoformer}
Haixu Wu, Jiehui Xu, Jianmin Wang, and Mingsheng Long.
\newblock Autoformer: Decomposition transformers with auto-correlation for
  long-term series forecasting.
\newblock {\em Advances in Neural Information Processing Systems},
  34:22419--22430, 2021.

\bibitem{zeng2023transformers}
Ailing Zeng, Muxi Chen, Lei Zhang, and Qiang Xu.
\newblock Are transformers effective for time series forecasting?
\newblock In {\em Proceedings of the AAAI conference on artificial
  intelligence}, volume~37, pages 11121--11128, 2023.

\bibitem{zhang2022less}
Tianping Zhang, Yizhuo Zhang, Wei Cao, Jiang Bian, Xiaohan Yi, Shun Zheng, and
  Jian Li.
\newblock Less is more: Fast multivariate time series forecasting with light
  sampling-oriented mlp structures.
\newblock {\em arXiv preprint arXiv:2207.01186}, 2022.

\bibitem{zhang2022crossformer}
Yunhao Zhang and Junchi Yan.
\newblock Crossformer: Transformer utilizing cross-dimension dependency for
  multivariate time series forecasting.
\newblock In {\em The Eleventh International Conference on Learning
  Representations}, 2022.

\bibitem{zhou2021informer}
Haoyi Zhou, Shanghang Zhang, Jieqi Peng, Shuai Zhang, Jianxin Li, Hui Xiong,
  and Wancai Zhang.
\newblock Informer: Beyond efficient transformer for long sequence time-series
  forecasting.
\newblock In {\em Proceedings of the AAAI conference on artificial
  intelligence}, volume~35, pages 11106--11115, 2021.

\bibitem{Zhou_Zhang_Peng_Zhang_Li_Xiong_Zhang_2022}
Haoyi Zhou, Shanghang Zhang, Jieqi Peng, Shuai Zhang, Jianxin Li, Hui Xiong,
  and Wancai Zhang.
\newblock Informer: Beyond efficient transformer for long sequence time-series
  forecasting.
\newblock {\em Proceedings of the AAAI Conference on Artificial Intelligence},
  page 11106–11115, Sep 2022.

\bibitem{zhou2022fedformer}
Tian Zhou, Ziqing Ma, Qingsong Wen, Xue Wang, Liang Sun, and Rong Jin.
\newblock Fedformer: Frequency enhanced decomposed transformer for long-term
  series forecasting.
\newblock In {\em International Conference on Machine Learning}, pages
  27268--27286. PMLR, 2022.

\end{thebibliography}

\end{document}